\documentclass[10pt,twocolumn]{article}

\usepackage{iccv}
\makeatletter
\@namedef{ver@everyshi.sty}{}
\makeatother
\usepackage{cite}
\usepackage{floatrow}
\newfloatcommand{capbtabbox}{table}[][\FBwidth]
\usepackage{tikz}
\usepackage{times}
\usepackage{epsfig}
\usepackage{graphicx}
\usepackage{color}
\usepackage{amsmath}
\usepackage{amssymb}
\usepackage{pifont}%
\usepackage{booktabs}
\usepackage{siunitx}

\usepackage{pbox}
\usepackage{tabularx}
\usepackage{todonotes}
\usepackage{multirow}
\usepackage{multicol}
\usepackage{float}
\usepackage[normalem]{ulem}
\useunder{\uline}{\ul}{}

\usepackage{makecell}

\definecolor{rev}{rgb}{0,0.0,0}
\newcommand\rev[1]{\textcolor{rev}{#1}}
\newcommand\revision[1]{\textcolor{rev}{#1}}

\definecolor{antonio}{rgb}{1.0,0.0,0.0}

\definecolor{kosta}{rgb}{0.0,0.0,1.0}

\definecolor{daniel}{rgb}{0.0,1.0,0.0}

\usepackage[pagebackref=true,breaklinks=true,letterpaper=true,colorlinks,bookmarks=false]{hyperref}

\iccvfinalcopy %
\def\iccvPaperID{1773} %

\ificcvfinal\fi

\definecolor{somegray}{rgb}{0.5, 0.5, 0.5}
\newcommand{\darkgrayed}[1]{\textcolor{somegray}{#1}}
\makeatletter
\newcommand*\titleheader[1]{\gdef\@titleheader{#1}}
\AtBeginDocument{%
  \let\st@red@title\@title
  \def\@title{%
    \vskip-3em
    \bgroup\normalfont\large\centering\@titleheader\par\egroup
    \vskip1.5em\st@red@title}
}
\makeatother

\titleheader{\darkgrayed{This paper has been accepted for publication at the\\
IEEE International Conference on Computer Vision (ICCV), Seoul, 2019.
\copyright IEEE}}

\title{End-to-End Learning of Representations for Asynchronous Event-Based Data}

\begin{document}

\author{Daniel Gehrig$^1$, Antonio Loquercio$^1$, Konstantinos G. Derpanis$^2$, Davide Scaramuzza$^1$\\
$^1$ Robotics and Perception Group\\
Depts. Informatics and Neuroinformatics\\
University of Zurich and ETH Zurich\\
$^2$  Ryerson University and Samsung AI Centre Toronto
}

\maketitle
\thispagestyle{empty}

\vspace{-1ex}
\begin{abstract}
   Event cameras are vision sensors that record asynchronous streams of per-pixel brightness changes, referred to as ``events''.
   They have appealing advantages over frame-based cameras for computer vision, including high temporal resolution, high dynamic range, and no motion blur.
   Due to the sparse, non-uniform spatiotemporal layout
   of the event signal, pattern recognition algorithms typically aggregate events %
   into a grid-based representation and subsequently 
   process it
   by a standard vision pipeline, e.g., Convolutional Neural Network (CNN).
   In this work, we introduce a general framework to convert event streams into grid-based representations through a sequence of differentiable operations. 
   Our framework comes with two main advantages: (i) allows learning the input event representation together with the task dedicated network in an end-to-end manner, and (ii) lays out a taxonomy that unifies the majority of extant event  representations in the literature and identifies novel ones.
Empirically, we show that our approach to learning the event representation end-to-end yields an improvement of approximately 12\% on optical flow estimation and object recognition over state-of-the-art methods.

\end{abstract}
\vspace{-1ex}
\section*{Multimedia Material}
The project's code is available on the following page: \url{https://github.com/uzh-rpg/rpg_event_representation_learning}. Additionally, qualitative results can be viewed in this video: \url{https://youtu.be/bQtSx59GXRY}
\section{Introduction}
\begin{figure}[t]
    \centering
    \includegraphics[height=3.2in]{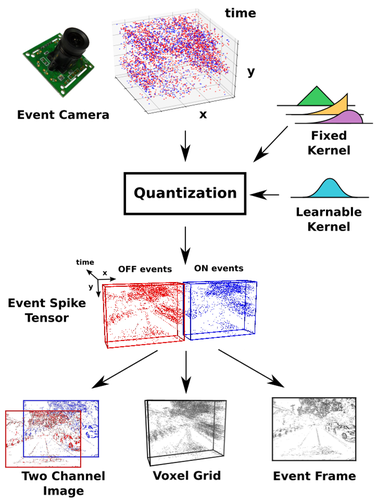}\\
    \vspace{-1ex}
     \caption{General framework to convert asynchronous event data  into grid-based  representations  using convolutions, quantization, and projections.
     All of these operations are differentiable. Best viewed in color.\vspace{-3.5ex}}
    \label{fig:cars_classification}
\end{figure}

Event cameras are bio-inspired vision sensors that operate radically differently from traditional cameras.
Instead of capturing brightness images at a fixed rate, event cameras measure brightness \emph{changes} (called \emph{events})
for each pixel independently.
Event cameras, such as the Dynamic Vision Sensor (DVS)~\cite{Lichtsteiner08ssc}, possess appealing properties compared to traditional frame-based cameras, including
a very high dynamic range, high temporal resolution (in the order of %
microseconds), and low power consumption.
In addition, event cameras greatly reduce bandwidth.
While frame-based cameras with comparable temporal resolution and/or dynamic range cameras exist, they are typically bulky, power-intensive, and require cooling \cite{phantomcamera}.%

The output of an event camera consists of a stream of events that encode the time, location, and polarity (sign) of the brightness changes.
Consequently, each event alone carries very little information about the scene.
Event-based vision algorithms aggregate information to enable further processing %
in two ways:
(i) use a continuous-time model (\emph{e.g.}, Kalman filter) that can be updated asynchronously with each incoming event~\cite{Mueggler14iros,Orchard15pami,Lagorce16pami,Gallego17pami,Andreopoulos18cvpr}
or (ii) process events simultaneously in packets ~\cite{Kim16eccv,Zhu17cvpr,Rebecq17ijcv,Zhu18eccv,Rosinol18ral}, \emph{i.e}., spatiotemporal localized aggregates of events.
The former methods can achieve minimal latency, but are sensitive to parameter tuning (\emph{e.g.}, filter weights) and are computationally intensive, since they perform an update step for each event.
In contrast, methods operating on event packets trade-off latency for computational efficiency and performance.
Despite their differences, both paradigms have been successfully applied on various vision tasks, including 
tracking~\cite{Gehrig18eccv,Gallego17pami,Mueggler14iros,Ni12jm},
depth estimation~\cite{Andreopoulos18cvpr,Rebecq17ijcv,Zhu18eccv}, 
visual odometry~\cite{Kim16eccv,Rebecq17ral,Zhu17cvpr,Rosinol18ral},
recognition~\cite{Orchard15pami,Lagorce16pami}, and optical flow estimation~\cite{Benosman14tnnls,Zhu18rss}.
A good survey on the applications of event cameras can be found in \cite{Gallego19arxiv}.
Motivated by the broad success of deep learning in computer vision on frame-based 
imagery, a growing number of recent event-based works have adopted a
data driven approach %
\cite{PerezCarrasco13pami,Lee16fns, Amir17cvpr, Maqueda18cvpr,Zhu18rss}.
Spiking Neural Networks (SNNs) are a natural fit to process 
event streams,
since they enable asynchronous inference at low power on specialized hardware~\cite{PerezCarrasco13pami,Lee16fns,Amir17cvpr}.
However, SNNs are notoriously difficult to train, as no efficient backpropagation algorithm exists~\cite{huh2018gradient}.
In addition, the special-purpose hardware required to run SNNs is expensive and in the development stage, which hinders its widespread adoption in the vision community.

Most closely related to the current paper are methods that pair an event stream with standard frame-based deep convolutional neural network (CNN) or recursive architectures, \emph{e.g.}, \cite{Sironi18cvpr,Lagorce16pami,neil2016phased,Maqueda18cvpr,Zhu18rss}.
To do so, a pre-processing step typically converts asynchronous event data to a grid-like representation, which can be updated either synchronously~\cite{Maqueda18cvpr,Zhu18rss} or asynchronously~\cite{Sironi18cvpr,Lagorce16pami}.
These methods benefit from their ease of implementation using standard
frame-based deep learning libraries (\emph{e.g.}, \cite{Abadi2016,paszke2017}) and fast inference on commodity graphics hardware.
However, these efforts have mainly focused on the downstream task beyond the initial representational stage and simply consider a fixed, possibly suboptimal, conversion between the raw event stream and the input grid-based tensor.
To date, there has not been an extensive study on the impact of the choice of input representation, leaving the following fundamental open question:
\rev{\emph{What is the best way to convert an asynchronous event stream into a grid-based (tensor) representation to maximize the performance on a given task?}}
In this paper, we aim to address this knowledge gap.

\noindent{\bf Contributions } We propose a general framework that converts asynchronous event-based data into grid-based representations. 
To achieve this, we express the conversion process through kernel convolutions, quantizations, and projections, where each operation is differentiable (see Fig.\ \ref{fig:cars_classification}).
Our framework comes with two main advantages.
First, it makes the conversion process fully differentiable, allowing to learn a representation \emph{end-to-end} from raw event data to the task loss.
In contrast, prior work assumes the input event representation as fixed.
Second, it lays out a taxonomy that unifies the majority of extant event 
representations in the literature and identifies novel ones.
Through extensive empirical evaluations we show that our approach to
learning the event representation end-to-end yields an improvement of 12\% on optical flow and 12.6\% on object recognition over state-of-the-art approaches that rely on handcrafted input event representations. %
In addition, we compare our methodology to asynchronous approaches in term of accuracy and computational load to shed light on the relative merits of each category.

\section{Related Work}

Traditionally,  
handcrafted features were used in frame-based computer vision, \emph{e.g.}, \cite{Lowe04ijcv, Sivic09pami, Leutenegger11iccv, Viola04ijcv, Dollar12pami}. %
More recently, research has shifted towards data-driven models, where features are automatically learned from data, \emph{e.g.}, ~\cite{He16cvpr, Redmon16cvpr,Arandjelovic16cvpr, Ilg17cvpr, Meister18aaai}.
The main catalyst behind this paradigm shift has been the availability of large training datasets~\cite{Deng09cvpr, Everingham09ijcv, Dosovitskiy15iccv}, efficient learning algorithms~\cite{le11icml,srivastava14jmlr} and suitable hardware.
Only recently has event-based vision made strides to address each of these areas.

Analogous to early frame-based computer vision approaches, significant effort has been made in designing efficient spatiotemporal feature descriptors of the event stream. 
From this line of research, typical high-level applications are gesture recognition~\cite{Lee12iscas}, object recognition~\cite{Sironi18cvpr, Lagorce16pami, park2015icip} or face detection~\cite{Barua16wacv}.
Low-level applications include optical flow prediction~\cite{Benosman11tnn, Benosman12nn} and image reconstruction~\cite{Bardow16cvpr}.

Another line of research has focused on applying data-driven models to event-based data.  
\revision{These include asynchronous, spiking neural networks (SNNs)\footnote{\rev{Here we use the term SNNs as in the neuromorphic literature~\cite{Lee16fns}, where it describes continuous-time neural networks.
Other networks which are sometimes called SNNs are low precision networks, such as binary networks \cite{Rastegari2016eccv}. However, these are not well suited for asynchronous inputs.}}~\cite{Lee16fns}, which have been  applied to several tasks, \emph{e.g.}, object recognition~\cite{Orchard15pami, Zhao2015nnls, Lee16fns, PerezCarrasco13pami},  gesture classification~\cite{Amir17cvpr}, and
optical flow prediction~\cite{Benosman14tnnls, Benosman12nn}.} 
However, the lack of specialized hardware and computationally efficient backpropagation algorithms still limits the usability of SNNs in complex real-world scenarios.
A typical solution to this problem is learning parameters with frame-based data and transferring the learned parameters to event data~\cite{PerezCarrasco13pami, Diehl15ijcnn}.
However, it is not clear how much this solution can generalize to real, noisy, event data that has not been observed during training.

\begin{table*}[t!]
\centering
\footnotesize

\begin{tabular}{l|lll}\hline

\textbf{Representation} & \textbf{Dimensions} & \textbf{Description} & \textbf{Characteristics}\\\hline
Event frame~\cite{Rebecq17bmvc} & $H \times W$  & {Image of event polarities} &Discards temporal and polarity information\\
Event count image~\cite{Maqueda18cvpr, Zhu18rss}& $2\times H \times W$ & Image of event counts & Discards time stamps\\
Surface of Active Events (SAE)~\cite{Benosman14tnnls,Zhu18rss}& $2 \times H \times W$ & Image of most recent time stamp & Discards earlier time stamps\\
Voxel grid~\cite{Zhu19cvpr}& $B \times H \times W$ & Voxel grid summing event polarities & Discards event polarity\\
Histogram of Time Surfaces (HATS)~\cite{Sironi18cvpr}& $2 \times H \times W$  & Histogram of average time surfaces & Discards temporal information\\
Event Spike Tensor (EST, our work) & $2 \times B \times H \times W$  & Sample event point-set into a grid & Discards the least amount of information \\\hline
\end{tabular}
\caption{Comparison of grid-based event representations used in prior work on event-based deep learning. $H$ and $W$ denote the image height and width dimensions,
respectively, and $B$ the number of temporal bins.
\vspace{-3ex}}
\label{tab:relatedwork}
\end{table*}

Recently, several works have proposed to use standard learning architectures as an alternative to SNNs~\cite{Maqueda18cvpr, neil2016phased, Zhu18rss, Zhu18ral, Sironi18cvpr}.
To process asynchronous event streams, Neil et al.~\cite{neil2016phased} adapted a recursive architecture to include the time dimension for prediction.
Despite operating asynchronously, their approach introduces high latency, since events have to pass sequentially through the entire recursive structure.
To reduce latency, other methods convert event streams into a grid-based representation, compatible with learning algorithms designed for standard frames, \emph{e.g.}, CNNs~\cite{Maqueda18cvpr, Zhu18rss, Zhu18ral, Sironi18cvpr}.
Sironi et al.~\cite{Sironi18cvpr} obtained state-of-the-art results in object recognition tasks by transforming events into histograms of averaged time surfaces (HATS), which are then fed to a support vector machine for inference.
The main advantage of their representation is that it can not only be used in conjunction with standard learning pipelines, but it can also be updated asynchronously, if sufficient compute is available.
A simpler representation was proposed by Maqueda et al.~\cite{Maqueda18cvpr} to address steering-angle prediction, where events of different polarities are accumulated over a constant temporal window.
To perform a low-level task, \emph{i.e.}, optical flow estimation, Zhu et al.\cite{Zhu18rss} proposed to convert events into a four-dimensional grid that includes both the polarity and spike time.
Finally, Zhu et al.\cite{Zhu19cvpr} converted events into a spatiotemporal voxel-grid.
Compared to the representation proposed in~\cite{Maqueda18cvpr}, the two latter representations have the advantage of preserving temporal information.
A common aspect among these works is the use of a 
handcrafted event stream representation.
In contrast, in this paper we propose a novel
event-based representation that is learned end-to-end
together with the task.
A comparison of event-based representations
and their design choices is summarized 
in Table~\ref{tab:relatedwork}.

Coupling event-based data with standard frame-based learning architectures has the potential to realize the flexibility of learning algorithms with the advantages of event cameras.
It is however not yet clear what is the impact of the event representation on the task performance. %
In this work, we present an extensive empirical study on the choice 
of representation for the 
the tasks of object recognition and optical flow estimation, central tasks in computer vision.

\newcommand{\eventMapping}{\mathcal{M}}
\newcommand{\eventTensor}{\mathcal{T}}
\newcommand{\uEv}{\textbf{u}}
\newcommand{\Brightness}[1]{L(x,y,#1)}
\newcommand{\eventSet}{\mathcal{E}}
\newcommand{\dirac}[1]{\delta(#1)}
\newcommand{\meas}{\textbf{m}}
\newcommand{\kernel}{\textbf{k}}

\vspace{-1ex}
\section{Method}
\begin{figure}[t]
    \centering
    \includegraphics[width=3in]{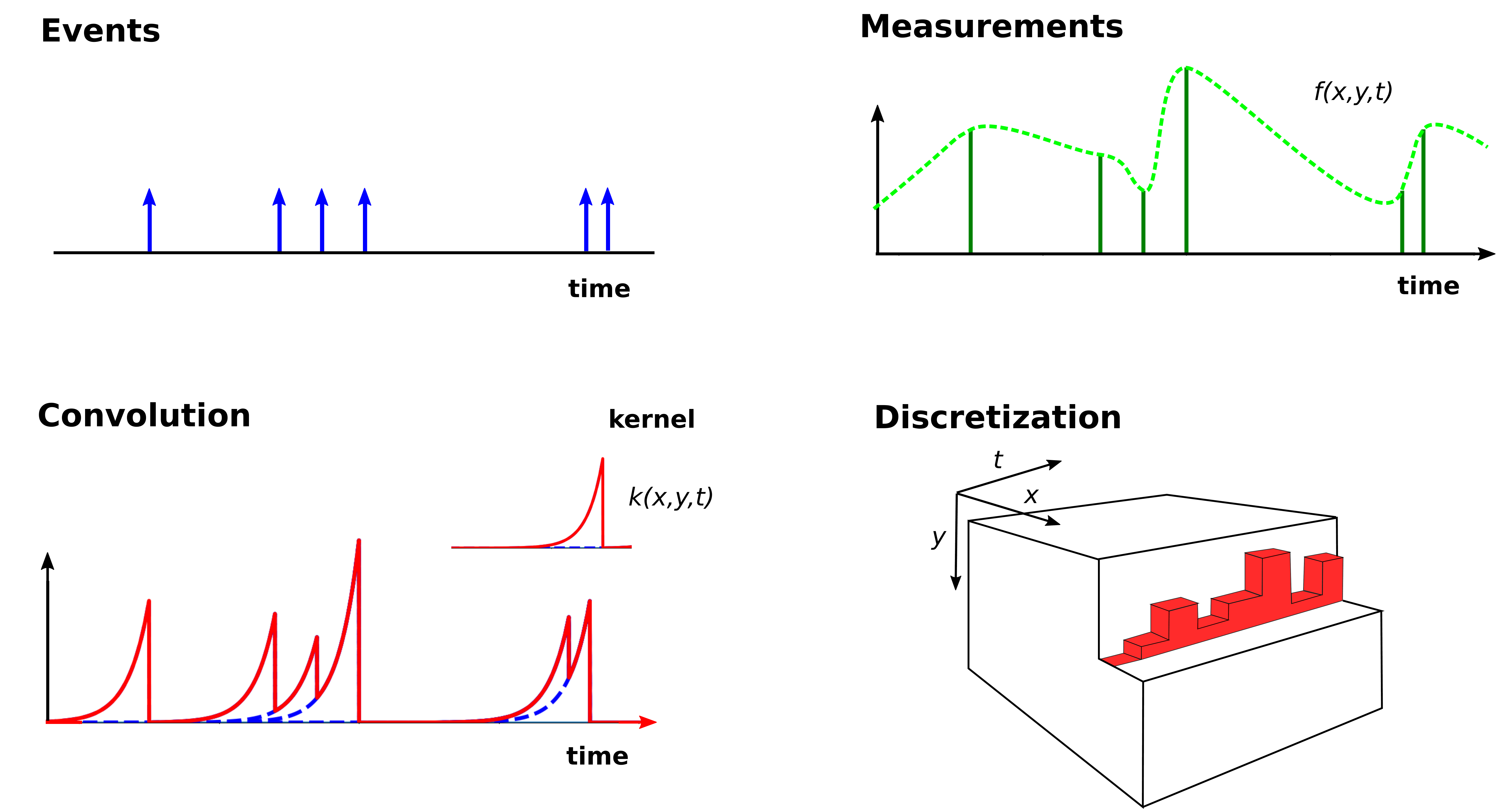}
    \caption{An overview of our proposed framework.
    Each event is associated with a measurement (green) which is convolved with a (possibly learnt) kernel. 
    This convolved signal is then sampled on a regular grid. Finally, various representations can be instantiated by performing projections over the temporal axis or over polarities.\vspace{-2ex}}
    \label{fig:explanation_method}
\end{figure}

\vspace{-1ex}
In this section, we present a general framework to convert asynchronous event streams into grid-based representations.
By performing the conversion strictly through differentiable operators, our framework allows us to learn a representation end-to-end for a given task.
Equipped with this tool, we derive a taxonomy that unifies common representations in the literature and identifies new ones.
An overview of the proposed framework is given in Fig.~\ref{fig:explanation_method}.

\subsection{Event Data}
Event cameras have pixels which trigger events independently whenever there is a log brightness change: 
\begin{equation}
    \Brightness{t}-\Brightness{t-\Delta t} \geq p C,
\end{equation}
where $C$ is the contrast threshold, $p\in\{-1,1\}$ is the polarity of the change in brightness, and $\Delta t$ is the time since the last event at $\uEv=(x,y)^\top$. 
In a given time interval $\Delta \tau$, the event camera will trigger a number of events: 
\begin{equation}
\label{eq:pointset}
    \eventSet = \{e_k\}_{k=1}^N = \{(x_k, y_k, t_k, p_k )\}_{k=1}^N.
\end{equation}
Due to their asynchronous nature, events are represented as a set.
To use events in combination with a convolutional neural network it is necessary to
convert the event set into a grid-like representation.
This means we must find a mapping $\eventMapping:\eventSet\mapsto\eventTensor$ between the set $\eventSet$ and a tensor $\eventTensor$.
Ideally, this mapping should preserve the structure (\textit{i.e.}, spatiotemporal locality) and information of the events.

\subsection{Event Field}
Intuitively, events represent point-sets in a four-dimensional manifold spanned by the $x$ and $y$ spatial coordinates, time, and polarity.
This point-set can be summarized by the \emph{event field}, inspired by~\cite{Censi2015acc,Lee16fns}:
\begin{equation}
  \label{eq:spike_train}
  S_{\pm}(x,y,t) = \sum_{e_k\in\eventSet_\pm} \dirac{x-x_k,y-y_k}\dirac{t-t_k},
\end{equation}
defined in continuous space and time, for events of positive ($\eventSet_+$) and negative ($\eventSet_-$) polarity.
This representation %
replaces each event by a Dirac pulse in the space-time manifold.
The resulting function $S_{\pm}(x,y,t)$ gives a continuous-time representation of $\eventSet$ which preserves the event's high temporal resolution and enforces spatiotemporal locality. 

\subsection{Generating Representations}
\label{sec:generating_representations}
\paragraph{Measurements}
In this section, we generalize the notion of the event field and demonstrate how it can be used to generate a grid-like representation from the events.
We observe that \eqref{eq:spike_train} can be interpreted as successive measurements of a function $f_\pm$ defined on the domain of the events, \textit{i.e.}, 
\begin{equation}
  \label{eq:spike_train_general}
  S_{\pm}(x,y,t) = \sum_{e_k\in\eventSet_\pm} f_\pm (x,y,t)\dirac{x-x_k,y-y_k}\dirac{t-t_k}.
\end{equation}
We call (\ref{eq:spike_train_general}) the \emph{Event Measurement Field}.
It assigns a measurement $f_\pm(x_k,y_k,t_k)$ to each event.
Examples of such functions are the event polarity $f_\pm(x,y, t)=\pm1$, the event count $f_\pm(x,y,t)=1$, and the normalized time stamp $f_\pm(x,y,t)=\frac{t-t_0}{\Delta t}$.
Other examples might include the instantaneous event rate or image intensity provided by such sensors as the Asynchronous Time-based Image Sensor (ATIS) \cite{Brandli14ssc}.
Various representations in the literature make use of the event measurement field.
In several works \cite{Sironi18cvpr, Lagorce16pami, Maqueda18cvpr, Zhu18rss}, pure event counts are measured, and summed for each pixel and polarity to generate \emph{event count images}.
Other works \cite{Zhu18rss, Benosman14tnnls} use the time stamps of the events to construct the \emph{surface of active events} (SAE) which retains the time stamp of the most recent event for each pixel and polarity.
Other representations use the event polarities and aggregate them into a three-dimensional \emph{Voxel Grid} \cite{Zhu19cvpr} or a two-dimensional \emph{Event Frame} \cite{Rebecq17bmvc}.

\vspace{-1ex}
\paragraph{Kernel Convolutions}
Although the event measurement field retains the high temporal resolution of the events, it is still ill-defined due to the use of Dirac pulses.
Therefore, to derive a meaningful signal from the event measurement field, we must convolve it with a suitable aggregation kernel.
The convolved signal thus becomes:
\begin{align}
  \label{eq:spike_train_general_convolved}
  &(k*S_{\pm})(x,y,t) \nonumber\\
  &=\sum_{e_k\in\eventSet_\pm} f_\pm (x_k, y_k, t_k)k(x-x_k,y-y_k,t-t_k).
\end{align}
In the literature, \eqref{eq:spike_train_general_convolved} is also known as the \emph{membrane potential}~\cite{Lee16fns,Ponulak2005ReSuMe,Mohemmed2012ijns}.
Several variations of this kernel have been used in prior works.
The two most commonly used ones are the $alpha$-kernel, $k(x,y,t) = \delta(x,y)\frac{e t}{\tau}\exp{(-t/\tau)}$ \cite{Lee16fns,Mohemmed2012ijns}, and the exponential kernel,  $k(x,y,t) = \delta(x,y)\frac{1}{\tau} \exp{(-t/\tau)}$ \cite{Ponulak2005ReSuMe}.
In fact, the exponential kernel is also used to construct the hierarchy of time-surfaces (HOTS) \cite{Lagorce16pami} and histogram of average time-surfaces (HATS) \cite{Sironi18cvpr}, where events are aggregated into \emph{exponential time surfaces}.
In the case of HATS \cite{Sironi18cvpr}, the exponential time surfaces can be interpreted as a local convolution of the spike train with an exponential kernel.
Another kernel which is typically used is the trilinear voting kernel, $k(x,y,t)=\delta(x,y)\max{(0,1-\vert \frac{t}{\Delta t}\vert)}$ \cite{Jaderberg15NIPS}.
Generally, the design of kernel functions is based on task-dependent heuristics with no general agreement on the optimal kernel to maximize task performance.

\vspace{-1ex}
\paragraph{Discretized Event Spike Tensor}
After kernel convolutions, a grid representation of events can be realized by sampling the convolved signal, \eqref{eq:spike_train_general_convolved}, at regular intervals:
\begin{align}
\label{eq:event_tensor}
     &S_\pm[x_l, y_m, t_n] = (k*S_\pm)(x_l,y_m t_n) \\
     &=  \sum_{e_k\in\eventSet_\pm} f_\pm (x_k,y_k, t_k)k(x_l-x_k,y_m-y_k,t_n-t_k).\nonumber
\end{align}
Typically, the spatiotemporal coordinates, $x_l, y_m, t_n$, lie on a voxel grid, \textit{i.e.}, $x_l \in \{0,1,...,W-1\}, y_m \in \{0,1,...,H-1\}$, and $t_n \in \{t_0,t_0+\Delta t,...,t_0 + B \Delta t\}$,  where $t_0$ is the first time stamp, $\Delta t$ is the bin size, and $B$ is the number of temporal bins.
We term this generalized representation the \emph{Event Spike Tensor} (EST).
Summing over both the polarity and time dimensions, one can derive the event-frame representation introduced in prior work~\cite{Rosinol18ral}.
Previous works considered quantizing various dimensions,
including spatiotemporal binning  \cite{Zhu19cvpr},
and quantizing both the polarity and spatial dimensions \cite{Maqueda18cvpr, Zhu18rss}.
However, the generalized form that retains all four dimensions has not been previously considered, and thus is a
\textit{new representation}.

\vspace{-2ex}
\paragraph{End-to-end Learned Representations}
The measurement and kernel in \eqref{eq:event_tensor} are generally hand crafted functions.
Previous works manually tuned those functions to maximize task performance.
In contrast, we propose to leverage the data directly to find the best function candidate, thus learning the representation \emph{end-to-end}.
We achieve this by replacing the kernel function in \eqref{eq:event_tensor} with a multilayer perceptron (MLP) with two hidden layers each with 30 units.
This MLP takes the coordinates and time stamp of an event as input, and produces an activation map around it.
For each grid location in the representation we evaluate the activation maps produced by each event and sum them together according to \eqref{eq:event_tensor}.
This operation is repeated for every point in the final grid, resulting in a grid-like representation.
To enforce symmetry across events, we limit the MLP input to the difference in coordinates $x_l-x_k, y_m-y_k, t_l-t_k$.
For the sake of simplicity, we do not learn the measurement function as well, choosing it instead from a set of fixed functions.
To speed up inference, at test time the learnt kernel can be substituted with an efficient look-up table, thus having comparable computation cost to handcrafted kernels.
These design choices make the representation both efficient and fully differentiable.
In contrast to previous works that used sub-optimal heuristics to convert events into grids, our framework can now tune the representation
to the downstream task, thus maximizing performance.

\vspace{-2ex}
\paragraph{Projection}
From the generalized \textit{event spike tensor} we can further instantiate novel and existing representations.
Many works for example deal with three-dimensional tensors, such as \cite{Sironi18cvpr, Lagorce16pami,Maqueda18cvpr,Zhu18rss,Zhu19cvpr}.
The event spike tensor, being a four-dimensional data structure \rev{(two spatial, one temporal, and one polarity)}, thus acts as a precursor to these three-dimensional constructs, which can be obtained by summing over one of the four dimensions.
For example, the Two-Channel Image \cite{Maqueda18cvpr, Sironi18cvpr,Zhu18rss}, can be derived by contracting the temporal dimension, either through summation \cite{Maqueda18cvpr, Sironi18cvpr, Zhu18rss} or maximization \cite{Zhu18rss}.
The voxel grid representation~\cite{Zhu19cvpr} can be derived by summing across event polarities.
All of these operations can be generalized via the projection operator $H_v$, where $H$ can be summation $\Sigma$, maximization $\max$, \etc and $v$ denoting the dimension can be $x_l$, $y_m$, $t_n$, or over polarity $\pm$, yielding 16 possible projections.
Here, we list only the representations that retain the spatial dimension, of which there are four, including the EST without projection:
\begin{align}
\multicolumn{2}{c}{$S_\pm[x_l,y_m, t_n]$}\label{eq:EST} \\
     S[x_l,y_m, t_n] &= H_\pm(S_\pm[x_l, y_m, t_n])\label{eq:Voxel} \\
     S_\pm[x_l,y_m] &= H_{t_n}(S_\pm[x_l, y_m, t_n]) \label{eq:channel}\\
     S[x_l,y_m] &= H_{t_n,\pm}(S_\pm[x_l, y_m, t_n]).\label{eq:Event}
\end{align}
We refer to these representations as the \emph{EST} \eqref{eq:EST}, \emph{Voxel Grid} \eqref{eq:Voxel}, \emph{Two-Channel Image} \eqref{eq:channel}, and \emph{Event Frame} \eqref{eq:Event}. 
The direction of projection has an impact on the information content of the resulting representation.
For example, projecting along the temporal axis greatly compresses the event representation, but at the cost of temporal localization information.
In contrast, projecting the event polarities leads to the cancellation of positive and negative events, potentially removing information in the process.
Of these representations, the EST stands out, as it retains the maximum amount of event information %
by forgoing the projection operation.

\section{Empirical Evaluation}
In this section, we present an extensive comparative 
evaluation of the representations identified by our taxonomy for object recognition (Sec.\ \ref{sec:object_recognition}) and optical flow estimation (Sec.\ \ref{sec:optical_flow}) on
standard event camera benchmarks.
\vspace{-1ex}
\paragraph{Candidate Representations}\label{sec:representations}
We start out by identifying 12 distinct representations based on the event spike tensor \eqref{eq:event_tensor}.
In particular, we select the measurement function \eqref{eq:spike_train_general} from three candidates: event polarity,  event count, and  normalized time stamp.
We use the summation operator $\Sigma$ to project out various axes defined in %
\eqref{eq:EST} - \eqref{eq:Event},  resulting in four variations: Event Spike Tensor, Voxel Grid, Two-Channel Image, and Event Frame.
We split the event spike tensor (a four-dimensional tensor) along the polarity dimension and concatenate the two tensors along the temporal dimension, effectively doubling the number of channels.
This is done to make the representation compatible with two-dimensional convolutions. 
As a first step we apply a generic trilinear kernel to convolve the event spike signal, and later study the effect of different kernels on performance when applied to the EST.
Finally, we report results for our end-to-end trained variant that directly utilizes raw events.

\subsection{Object Recognition}\label{sec:object_recognition}
Object recognition with conventional cameras remains challenging due to their low dynamic range, high latency, and tendency to motion blur.
In recent years, event-based classification has grown in popularity because it can address all these challenges. 

In this section, we investigate the performance of the event representations proposed in Sec.~\ref{sec:representations} on the task of event-based object recognition.
In particular, we aim to determine the relationship between the information content of the representation and classification accuracy.
We show that our end-to-end learned representation significantly outperforms the state-of-the-art \cite{Sironi18cvpr}.
We use two publicly available datasets in our evaluation: N-Cars \cite{Sironi18cvpr} (Neuromorphic-Cars) and N-Caltech101~\cite{Orchard15fns}.
N-Cars provides a benchmark for the binary task of car recognition in a scene. 
It contains $24,029$ event samples of $100$ ms length recorded by the ATIS event camera \cite{Posch11ssc}.
N-Caltech101 (Neuromorphic-Caltech101) is the event-based version of the popular Caltech101 dataset \cite{Fei06tpami}, and poses the task of multiclass recognition for event cameras. 
It contains $8,246$ samples and $100$ classes, which were recorded by placing an event camera on a motor and moving it in front of a screen projecting various samples from Caltech101.
\vspace{-2ex}
\paragraph{Implementation}
We use a ResNet-34 architecture \cite{He16cvpr} for each dataset.  The network is pretrained on color RGB images from ImageNet \cite{Russakovsky15ijcv}. %
To account for the different number of input channels and output classes between the pre-trained model and ours,
following approach~\cite{Maqueda18cvpr}: we replace the first and last layer of the pre-trained model with random weights and then finetune all weights on the task.
We train by optimizing the cross-entropy loss and use the ADAM  optimizer \cite{Kingma15iclr} with an initial learning rate of $1e-5$, which we reduce by a factor of two every $10,000$ iterations.
We use a batch-size of $60$ and $100$ for N-Caltech101 and N-Cars, respectively.

\vspace{-2ex}
\paragraph{Results}
\begin{table}[t!]
\small\addtolength{\tabcolsep}{-2pt}
\centering
\scalebox{0.7}{
\begin{tabular}{c|c|c|c|c}
\hline
\textbf{Representation} & \textbf{Measurement}         & \textbf{Kernel}            & \textbf{N-Cars} & \textbf{N-Caltech101} \\
\hline
Event Frame             &\multirow{4}{*}{polarity}    & \multirow{4}{*}{trilinear}& 0.866           & 0.587                 \\
Two-Channel Image       &                              &                            & 0.830           & 0.711                 \\
Voxel Grid              &                              &                            & 0.865           & 0.785                 \\
\textbf{EST (Ours)}              & & & 0.868           & 0.789                 \\\hline
Event Frame             &\multirow{4}{*}{count}       & \multirow{4}{*}{trilinear}& 0.799           & 0.689                 \\
Two-Channel Image       &                              &                            & 0.861           & 0.713                 \\
Voxel Grid              &                              &                            & 0.827           & 0.756                 \\
\textbf{EST (Ours)}              & & & 0.863           & 0.784                 \\\hline
Event Frame             &\multirow{4}{*}{time stamps} & \multirow{4}{*}{trilinear}& 0.890           & 0.690                 \\
Two-Channel Image       &                              &                            & 0.917           & 0.731                 \\
Voxel Grid              &                              &                            & 0.847           & 0.754                 \\
\textbf{EST (Ours)}              & & & 0.917           & 0.787                 \\\hline
\multirow{3}{*}{\textbf{EST (Ours)}}    & \multirow{3}{*}{time stamps}  & alpha                      & 0.911           & 0.739                 \\
                        &                              & exponential                & 0.909           & 0.782                 \\
                        &                              & learnt                     & \textbf{0.925}           & \textbf{0.817}                 \\ \hline
\end{tabular}}
\vspace{-5pt}
\caption{Classification accuracy for all event representations using different measurement functions, as described in Sec. \ref{sec:representations}. \rev{For each representation the temporal dimension was discretized into nine bins.} For the best performing representation (EST and time stamp measurements) we additionally report results for different kernel choices: trilinear \cite{Jaderberg15NIPS}, exponential \cite{Ponulak2005ReSuMe}, alpha kernels \cite{Lee16fns}, as well as a learnable kernel.}
\label{tab:classification_representations}
\end{table}
\begin{table}[t]
\small\addtolength{\tabcolsep}{-2pt}
\centering
\scalebox{0.7}{
\begin{tabular}{c|c|c|c|c}
\hline
\textbf{Representation}              & \textbf{Measurement}        & \textbf{Kernel}            & \textbf{N-Cars} & \textbf{N-Caltech101} \\ \hline
H-First\cite{Orchard15pami} &\multirow{5}{*}{-}&\multirow{5}{*}{-}& 0.561 & 0.054\\
HOTS\cite{Lagorce16pami}    &&& 0.624 & 0.210\\
Gabor-SNN\cite{Sironi18cvpr}&&& 0.789 & 0.196\\
HATS\cite{Sironi18cvpr}     &&& 0.902 & 0.642\\
HATS + ResNet-34             &&& 0.909 & 0.691\\\hline
Two-Channel Image\cite{Maqueda18cvpr}& count & \multirow{2}{*}{trilinear} & 0.861 & 0.713 \\
Voxel Grid\cite{Zhu19cvpr} & polarity &  & 0.865 & 0.785\\ \hline
\multirow{2}{*}{\textbf{EST (Ours)}} & \multirow{2}{*}{time stamps} & trilinear                  & 0.917           & 0.787                 \\
                                     &                             & learnt                     & \textbf{0.925}  & \textbf{0.817}        \\ \hline
\end{tabular}}
\vspace{-5pt}
\caption{Comparison of the classification accuracy for different baseline representations \cite{Maqueda18cvpr, Zhu19cvpr} and state-of-the-art classification methods \cite{Sironi18cvpr, Lagorce16pami, Orchard15pami}. As an additional baseline we pair the best performing representation from previous work (HATS \cite{Sironi18cvpr}) with a more powerful classification model (ResNet-34, used in this work) as the original numbers were reported using a linear SVM.\vspace{-3ex}}
\label{tab:classification_related_work}
\end{table}

The classification results are shown in Table \ref{tab:classification_representations}.
From the representations that we evaluated, the event spike tensor with time stamp measurements has the highest accuracy on the test set for both N-Cars and N-Caltech101.
From these results we can make two conclusions. First, we observe that representations that separate polarity consistently outperform those that sum over polarities. 
Indeed, this trend is observed for all measurement functions: discarding the polarity information leads to a decrease in accuracy of up to $7$\%.
Second, we see that representations that retain the temporal localization of events, \emph{i.e.}, the Voxel Grid and EST, consistently outperform their counterparts, which sum over the temporal dimension.
These observations indicate that both polarity and temporal information are important for object classification.
This trend explains why the EST leads to the most accurate predictions: it retains the maximum amount of information with respect to the raw event data.

Interestingly, using event time stamps as measurements is more beneficial than other measurements, since the information about polarity and event count is already encoded in the event spike tensor.
Indeed, using the time stamps explicitly in the tensor partially recovers the high temporal resolution, which was lost during the convolution and discretization steps of the event field.
We thus established that the EST with time stamp measurements performs best for object classification.
However, the effect of the temporal kernel
remains to be explored. 
For this purpose we experimented with the kernels described in Sec.\ \ref{sec:generating_representations}, namely the exponential \cite{Ponulak2005ReSuMe}, alpha \cite{Lee16fns}, and trilinear \cite{Jaderberg15NIPS} kernels. 
In addition, we evaluate our end-to-end trainable representation and report the results in Table \ref{tab:classification_representations}.
We see that using different handcrafted kernels %
negatively impacts the test accuracies.
In fact, applying these kernels to the event spikes decreases the effective temporal localization compared to the trilinear kernel by overlapping the event signals in the representation.
This makes it difficult for a network to learn efficiently how to identify individual events. 
Finally, we see that if we learn a kernel end-to-end we gain a significant boost in performance.
This is justified by the fact that the learnable layer finds an optimal way to draw the events on a grid, maximizing the discriminativeness of the representation.
\vspace{-2ex}
\paragraph{Comparison with State-of-the-Art}
We next compare our results with state-of-the-art object classification methods that utilize handcrafted event representations, such as HATS \cite{Sironi18cvpr}, HOTS \cite{Lagorce16pami}, as well as a baseline implementation of an SNN \cite{Sironi18cvpr}.
For the best performing representation (HATS) we additionally report the classification accuracies obtained with the same ResNet-34 used to evaluate the EST; the original work used a linear SVM.
Two additional baselines are used for comparison: 
(i) the histogram of events \cite{Maqueda18cvpr} (here Two-Channel Image), with event count measurements,  and (ii) the Voxel Grid~\cite{Zhu19cvpr} with polarity measurements. %

The results for these methods are summarized in Table \ref{tab:classification_related_work}.
Our method outperforms the state-of-the-art (HATS) and variant (HATS + ResNet-34), as well as the Voxel Grid and Two-Channel Image baselines by 2.3\%, 1.6\%, 6\% and 6.5\% on N-Cars and 17.5\%, 12.6\%, 3.2\% and 10.4\% on N-Caltech101, respectively.
In particular, we see that our representation is more suited for object classification than existing handcrafted features, such as HATS and HOTS, even if we use more complex classification models with these features.
This is likely due to HATS discarding temporal information, which, as we established, plays an important role in object classification.
It is important to note, compared to the state-of-the-art, our method does not operate asynchronously, or at low power with current hardware (as for example SNNs); however, we show in Sec.\ \ref{sec:computation} that our method can still operate at a very high framerate that is sufficient for many high-speed applications.
\subsection{Optical Flow Estimation}\label{sec:optical_flow}
Like object recognition, optical flow estimation using frame-based methods remains challenging in high-dynamic range scenarios, \emph{e.g.}, at night, and during high speed movements.
In particular, motion blur and over/under-saturation of the sensor often violate brightness constancy in the image, a fundamental assumption underlying many approaches, which leads to estimation errors.
Due to their lack of motion blur and high dynamic range, event cameras have the potential to provide higher accuracy estimates in these conditions.
Early works on event-based optical flow estimation fit planes to the spatiotemporal manifold generated by events \cite{Benosman14tnnls}.
Other works have tackled this task by finding the optimal event alignments when projected onto a frame \cite{Zhu17icra, Gallego18cvpr}.
Most recently, the relatively large-scale Multi Vehicle Stereo Event Camera Dataset (MVSEC) \cite{Zhu18ral} made possible deep learning-based optical flow \cite{Zhu19cvpr,Zhu18rss}. It provides data from a stereo DAVIS rig combined with a LIDAR for ground-truth optical flow estimation \cite{Zhu18rss}.
The dataset features several driving sequences during the day and night, and indoor sequences recorded onboard a quadcopter.
The methods in \cite{Zhu18rss,Zhu19cvpr} learn flow in a self-supervised manner and use standard U-Net architectures \cite{Ronneberger15icmicci}, outperforming existing frame-based methods in challenging night-time scenarios.
In \cite{Zhu18rss}, a four-channel image representation is used as input to the network.
This image is comprised of the two-channel event count image used in \cite{Maqueda18cvpr} and two-channel surface of active events (SAE) \cite{Benosman14tnnls},  divided according to event polarities.
While the event counts and time surfaces combine the temporal and spatial information of the event stream, it still compresses the event signal by discarding all event time stamps except the most recent ones.

To date, it is unclear which event representation is optimal to learn optical flow.
We investigate this question by comparing the representations listed in Sec.\ \ref{sec:representations} against the state-of-the-art \cite{Zhu18rss} for the task of optical flow regression, evaluated on the MVSEC dataset.
\vspace{-2ex}
\paragraph{Implementation}
We train an optical flow regressor on the outdoor sequences \emph{outdoor\_day1} and \emph{outdoor\_day2}.
These sequences are split into about $40,000$ samples at fixed time intervals. 
Each sample consists of events aggregated between two DAVIS frames, which are captured at $30$ Hz. 
We use EV-FlowNet \cite{Zhu18rss} as the base network, with the channel dimension of the initial convolution layer set to
the same number of channels 
of each input representation.
The network is trained from scratch using a supervised loss derived from ground truth motion field estimates:
\vspace{-1ex}
\begin{equation}
    l(f,f_\text{gt}) = \sum_x \rho(f-f_\text{gt}),
    \vspace{-1ex}
\end{equation}
where $\rho$ denotes the robust Charbonnier loss \cite{Sun2014ijcv}, $\rho(x) = (x^2+\epsilon^2)^\alpha$.
For our experiments, we chose $\epsilon=1e-3$ and $\alpha=0.5$. 
This loss is minimized using the ADAM optimizer \cite{Kingma15iclr} with an initial learning rate of $5e-5$ and reducing it by a factor of two after $40,000$ iterations and then again every $20,000$ iterations with a batch size of eight.
\vspace{-4ex}
\paragraph{Results}

\useunder{\uline}{\ul}{}
\begin{table*}[]
\centering
\scalebox{0.7}{\begin{tabular}{c|c|c|cc|cc|cc}
\hline
\multirow{2}{*}{\textbf{Representation}} & \multirow{2}{*}{\textbf{Measurement}} & \multirow{2}{*}{\textbf{Kernel}} & \multicolumn{2}{c|}{\textit{indoor\_flying1}} & \multicolumn{2}{c|}{\textit{indoor\_flying2}} & \multicolumn{2}{c}{\textit{indoor\_flying3}} \\ \cline{4-9} 
                                         &                                       &                                  & AEE                   & \% Outlier            & AEE                   & \% Outlier            & AEE                   & \% Outlier           \\ \hline
Two-Channel Image \cite{Maqueda18cvpr}                        & count                                 & \multirow{3}{*}{trilinear}                                 & 1.21                  & 4.49                  & 2.03                  & 22.8                  & 1.84                  & 17.7                 \\
EV-FlowNet \cite{Zhu18rss}                               & -                                     &        & 1.03                  & 2.20                  & 1.72                  & 15.1                  & 1.53                  & 11.9                 \\
Voxel Grid \cite{Zhu19cvpr}                               & polarity                              &                                  & 0.96                  & 1.47                  & 1.65                  & 14.6                  & 1.45                  & 11.4                 \\\hline
Event Frame                              &\multirow{3}{*}{time stamps}            & \multirow{3}{*}{trilinear}       & 1.17                  & 2.44                  & 1.93                  & 18.9                  & 1.74                  & 15.5                 \\
Two-Channel Image                        &                                       &                                  & 1.17                  & 1.5                   & 1.97                  & 14.9                  & 1.78                  & 11.7                 \\
Voxel Grid                               & && 0.98                  & 1.20                  & 1.70                  & 14.3                  & 1.5                   & 12.0                 \\\hline
\multirow{4}{*}{\textbf{EST (Ours)}}     & \multirow{4}{*}{time stamps}            & trilinear                        & 1.00                  & 1.35                  & 1.71                  & 11.4                  & 1.51                  & 8.29                 \\
                                         &                                       & alpha                            & 1.03                  & 1.34                  & 1.52                  & 11.7                  & 1.41                  & 8.32                 \\
                                         &                                       & exponential                      & \textbf{0.96}         & 1.27                  & 1.58                  & 10.5                  & \textbf{1.40}         & 9.44                 \\
                                         &                                       & learnt                           & 0.97                  & \textbf{0.91}         & \textbf{1.38}         & \textbf{8.20}         & 1.43                  & \textbf{6.47}        \\ \hline
\end{tabular}}
\vspace{-5pt}
\caption{Average end-point error (AEE) and \% of outliers evaluation on the MVSEC dataset for different variations of the EST with time stamp measurements.
\rev{For each representation the temporal dimension was discretized into nine bins.} 
Various baselines \cite{Maqueda18cvpr, Zhu19cvpr} and state-of-the-art methods \cite{Zhu18rss} are compared.\vspace{-3.5ex}} 
\label{tab:optic_flow_representations}
\end{table*}
As in \cite{Zhu18rss}, we measure the performance of our networks by comparing the average end-point error 
($\text{AEE} =  \frac{1}{N}\sum_i\left\vert f - f_\text{gt}\right\vert_2$)
on the \emph{indoor\_flying} datasets, which are visually distinct from the training set.
The test error on these datasets thus reflects the generalizability of our network, and its overall performance. 
In addition, as events only provide sparse information in the frame we only report the error computed at pixels where at least one event was triggered, as done in \cite{Zhu18rss}.
Following the KITTI 2015 benchmark \cite{Menze2015ISA}, we report the percentage of pixels which have an end-point-error larger than three pixels and 5\% of the ground-truth flow, also done in \cite{Zhu18rss}.
In the previous classification experiments we observed that time stamp measurements are essential for a discriminative representation.
We thus focus on results obtained from representations using the time stamp as the measurement function, as well as different kernels.
Table \ref{tab:optic_flow_representations} summarizes the results obtained from this experiment.
An exhaustive evaluation of the various measurement functions, \emph{i.e.}, polarities and counts, as well as qualitative results, is available in the supplemental material.

From Table \ref{tab:optic_flow_representations} we see that Voxel Grid and EST have similar AEE and outlier ratios.
This indicates that optical flow estimation is not as sensitive to event polarity as observed for classification.
This is further supported by the small performance gap between
the Two-Channel image and Event Frame.
A more striking difference comes when we compare representations which retain the temporal dimension (middle rows), with those that sum over it.
Indeed, the accuracies of the Two-Channel Image and the Event Frame drop approximately $10 - 20$\% when compared to the EST and Voxel-Grid.
As with the classification evaluation, we establish that EST is among the most competitive representations and further explore the influence of different kernels on the performance.
These are summarized in the bottom set of rows of Table \ref{tab:optic_flow_representations}.
We see that the exponential and alpha kernels outperform the trilinear kernel.
This indicates a strong dependency on the kernel shape and we thus proceed with the fully end-to-end learnable version.
As with classification, we observe that the learnable kernel significantly improves the accuracy on almost all scenes.
The most significant improvements are achieved for outlier ratios, indicating that using learnable kernels improves the robustness of the system.

\vspace{-2ex}
\paragraph{Comparison with State-of-the-Art}

We compare our method with 
the state-of-the-art
\cite{Zhu18rss}, as well as other baselines 
based on the representations used in \cite{Maqueda18cvpr} and \cite{Zhu19cvpr}.
Table \ref{tab:optic_flow_representations}
presents a detailed comparison.
It is clear that the EST outperforms the state-of-the-art by a large margin ($12$\%).
There are also significant improvements in terms of outlier ratio, reducing the outliers by an average of $49$\% which again indicates the robustness of our method.
This performance difference is likely due the data-driven nature of the learnable EST. 
While existing approaches learn the task on fixed event representations, our method learns the task and representation jointly. The resulting representation is more adapted to the task and thus maximizes performance.

\subsection{Computational Time and Latency}
\label{sec:computation}
\vspace{-2ex}
One of the key advantages of event cameras are their low latency and high update rate.
To achieve high-frequency predictions, previous works developed lightweight and fast algorithms to process each incoming event asynchronously.
In contrast, other approaches aggregate events into packets and then process them simultaneously.
While this sacrifices latency, it also leads to overall better accuracy, due to an increase in the signal-to-noise ratio.
Indeed, in several pattern recognition applications, \emph{e.g.}, object recognition and optical flow prediction, asynchronous processing is not essential: we may actually sacrifice it for improved accuracy.
We compare these two modes of operation in  Table \ref{tab:computation_time} where we show the number of events that can be processed per second, as well as the total time used to process a single sample of $100$ ms from the N-Cars dataset.
It can be seen that if we allow for batch computation, our method using a learnt kernel and lookup table can run at a very high speed that is comparable to other methods.
For applications where asynchronous updates or low-power consumption have higher priority than accuracy, other methods, \emph{e.g.}, SNNs, 
hold an advantage with respect to our approach.

\vspace{-1.5ex}
We further report the computation time per inference for different architectures in Table \ref{fig:inference_speed}.
We report the timing in two stages: representation computation and inference. 
While representation computation is performed on a CPU (Intel i7 CPU, 64bits, 2.7GHz and 16 GB of RAM), inference is performed on a GPU (GeForce RTX 2080 Ti).
Table \ref{fig:inference_speed} shows that the computation of the representation only contributes a small part to the overall computation time, while most of the time is spent during inference. 
Nonetheless, we see that a full forward pass only takes on the order of $6$ ms, which translates to a maximum inference rate of $146$ Hz. 
Although not on the order of the event rate, this value is high enough for most high-speed applications, such as mobile robotics or autonomous vehicle navigation.
Moreover, we see that we can reduce the inference time significantly if we use smaller models, achieving $255$ Hz for a ResNet-18. 
Shallower models could potentially be run at minimal loss in accuracy by leveraging distillation techniques~\cite{Hinton15NIPS}.

\begin{table}
\scalebox{0.7}{\begin{tabular}{c|c|c|c}
\hline
\textbf{Method} & \textbf{Asynchronous} & \textbf{Time [ms]} & \textbf{Speed {[}kEv/s{]}} \\ \hline
Gabor SNN   \cite{Sironi18cvpr} & Yes         & 285.95          & 14.15\\
HOTS \cite{Lagorce16pami}            & Yes  & 157.57                 & 25.68                                 \\
HATS \cite{Sironi18cvpr}           & Yes               & 7.28    & 555.74                                \\
\textbf{EST (Ours)}             & No                & \textbf{6.26}    & \textbf{632.9}                       \\ \hline
\end{tabular}}
\vspace{-5pt}
\caption{
Computation time for $100$ ms of event data and number of events processed per second. \vspace{-5pt}}
\label{tab:computation_time}
\end{table}

\begin{table}
\small\addtolength{\tabcolsep}{-2pt}
\scalebox{0.7}{\begin{tabular}{c|c|c|c|c}
\hline
\textbf{Model} & \textbf{Inference {[}ms{]}} & \textbf{Representation {[}ms{]}} & \textbf{Total {[}ms{]}} & \textbf{Rate {[}Hz{]}} \\ \hline
ResNet-18       & 3.87                        & 0.38                             & 4.25                    & 235                    \\
ResNet-34       & 6.47                        & 0.38                             & 6.85                    & 146                    \\
ResNet-50       & 9.14                        & 0.38                             & 9.52                    & 105                    \\
EV-FlowNet     & 5.70                        & 0.38                             & 6.08                    & 164                    \\ \hline
\end{tabular}}
\vspace{-5pt}
\caption{Computation time split into EST generation ($0.38$ ms) and inference for several standard network architectures. Both ResNet-34 \cite{He16cvpr} and EV-FlowNet \cite{Zhu18rss} allow processing at approximately $146$ Hz which is sufficient for most high-speed applications.\vspace{-3ex}}
\label{fig:inference_speed}
\end{table}

\section{Conclusions}

This paper presented a general framework for converting asynchronous event data into grid-based representations.
By representing the conversion process through differentiable operations, our framework allows learning input representations in a data-driven fashion.
In addition, our framework lays out a taxonomy which unifies a large number of extant event representations and identifies new ones.
Through an extensive evaluation we show that learning representations end-to-end together with the task yields an increase of about
$12$\%
in performance over state-of-the-art methods, for the tasks of object recognition and optical flow estimation.
With this contribution, we combined the benefits of deep learning with event cameras, thus unlocking their outstanding properties to a wider community.
As an interesting direction for future work, we plan to allow asynchronous updates by deploying recurrent architectures, similar to~\cite{neil2016phased}: this will bridge the gap between synchronous and asynchronous approaches for event-based processing.

\vspace{-1ex}
\section*{Acknowledgements}
\vspace{-1ex}
This project was funded by the Swiss National Center of Competence Research (NCCR) Robotics, through the Swiss National Science Foundation, and the SNSF-ERC starting grant. 
K.G.D. is supported by a Canadian NSERC Discovery grant. 
K.G.D. contributed to this work in his personal capacity as an Associate Professor at Ryerson University.
\section{Appendix}
We encourage the reader to watch the supplementary video at \url{https://www.youtube.com/watch?v=bQtSx59GXRY} for an introduction to the event camera and qualitative results of our approach.
In this section, we provide additional details about the network architecture used for our experiments, as well as supplementary results for object recognition and optical flow prediction.

\subsection{Network Architecture}

For all our classification experiments, we used an off-the-shelf ResNet-34 \cite{He16cvpr} architecture for inference with weights pretrained on RGB image-based ImageNet \cite{Russakovsky15ijcv}.
We then substitute the first and last layer of the pre-trained network with new weights (randomly initialized) to accommodate the difference in input channels (from the difference in representation) and output channels (for the difference in task).

For the optical flow experiments, we use the off-the-shelf U-Net architecture \cite{Ronneberger15icmicci} for inference, adapting its input layer to the number of channels of each representation.

\paragraph{Learned Kernel Functions}

\begin{figure}[t]
    \centering
    \includegraphics[width=4cm]{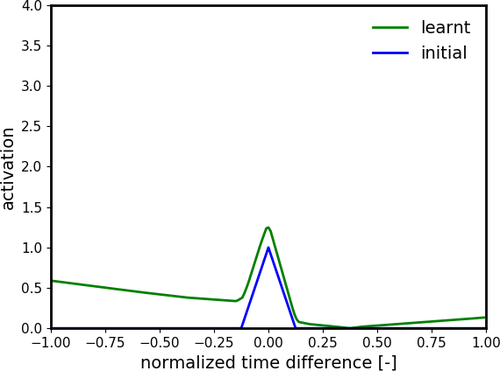}
    \includegraphics[width=4cm]{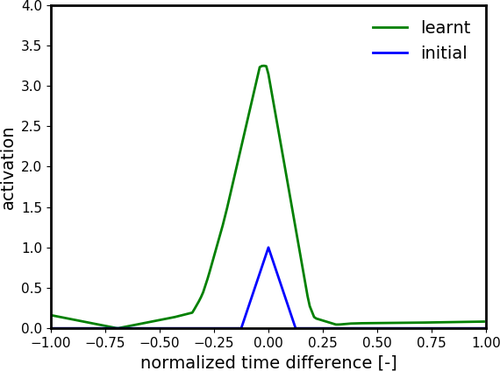}
    \caption{Kernel function learned for classification in the N-Cars dataset (left) and for optical flow prediction (right).}
    \label{fig:app:kernels}
\end{figure}

As discussed in Sec.~3.3 in the main manuscript, we used a two-layer multi-layer perceptron (MLP) to learn the kernel function to convolve the event measurement field, defined in \eqref{eq:spike_train_general}.
The two hidden layer have both $30$ nodes, with Leaky ReLU as activation function ($\text{leak}=0.1$) to encourage better gradient flow.
To give all image locations the same importance, we designed the kernel to be translation invariant.
Thus, for an event occurring at time $t_k$ the MLP has a one-dimensional input $\delta t = t^*_k - t_n$ and a single output $k(t^*_k-t_n)$ with normalized time $t^*_k = \frac{t_k}{\Delta t}$ and $\Delta t$ denoting the time window of the events.
The contribution of a single event to the sum in \eqref{eq:spike_train_general_convolved} is computed for every grid position $t_n$ for $n=0,1,...,B-1$ where $B$ is the number of temporal discretization bins.
The weights of the MLP were initialized with the trilinear voting kernel $k(x,y,t)=\delta(x,y)\max{(0,1-\vert \frac{t}{\Delta t}\vert)}$ \cite{Jaderberg15NIPS}, since this proved to facilitate convergence in our experiments.
Fig.~\ref{fig:app:kernels} shows an illustration of the learned kernels as a function of time.
Interestingly, the learned kernels show some interesting behavior, when compared against the trilinear voting kernel, on which they were initialized.
For classification (Fig.~\ref{fig:app:kernels}, left), the kernel seems to increase the event influence to the past, in a causual fashion: indeed, enough evidence has to be accumulated to produce a classification label.
In contrast, for optical flow prediction (Fig.~\ref{fig:app:kernels}, right), the learned kernel increases in magnitude, but not significantly in the time range, with respect to the trilinear kernel.
This is probably due to the fact that optical flow is a more `local' task with respect to classification, and therefore less temporal information is required.

\subsection{Ablation Studies and Qualitative Results}
\subsubsection{Classification}
\label{sec:app:classification}
\begin{figure*}[h]
    \centering
    \begin{tabular}{cc}
    	\textbf{Correct label: Car}&\textbf{Correct label: Car}\\ 
    	\includegraphics[height=1.7in]{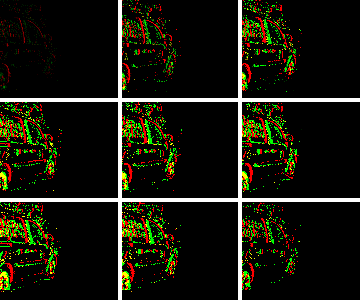}&
    	\includegraphics[height=1.7in]{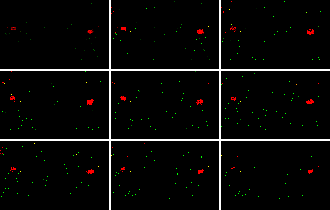}\\
    	good example: 99\% Car score&borderline example: 46\% Car score\\
    	(a)&(b)\\\\
    	\textbf{Correct label: Car}&\textbf{Correct Label: Car}\\
    	\includegraphics[height=1.7in]{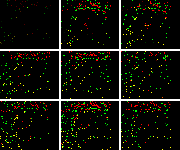}&
    	\includegraphics[height=1.7in]{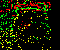}\\
    	bad example: 5\% Car score&improvement: 23\% Car score\\
    	(c)&(d)
    \end{tabular}
    \caption{Visualization of input representations derived from samples from the N-Cars dataset \cite{Sironi18cvpr} (a and b) show the event spike tensor (EST) representation with time measurements, which achieved the highest classification score on N-Cars, while (d) shows the two-channel image of sample (c) for comparison. The EST consists of 18 channels, where the first nine are filled with events of positive polarity and the last nine are filled with negative polarity. The images show the nine temporal bins of the tensor with positive events in red and negative events in green. 
    In good conditions (a) the classifier has high confidence in the car prediction. However, when there are less events due to the lack of motion (b and c) the uncertainty rises leading to predictions close to random (50\%). 
    In (b) the classifier sees the headlights of the car (red dots) but may still be unsure.
    In (c) the classifier sees only noise due to the high temporal resolution, likely attributing presence of noise to no motion.
    When we aggregate the noise (d) into the Two-Channel Image we see a more distinct pattern emerge, leading to higher classification confidence.  }
    \label{fig:app:cars_examples}
\end{figure*}

\begin{figure*}[h]
    \centering
    \begin{tabular}{ccc}
    	\textbf{Correct label: Butterfly}&\textbf{Correct label: Umbrella}&\textbf{Correct label: Strawberry} \\ 
    	\includegraphics[height=1.6in]{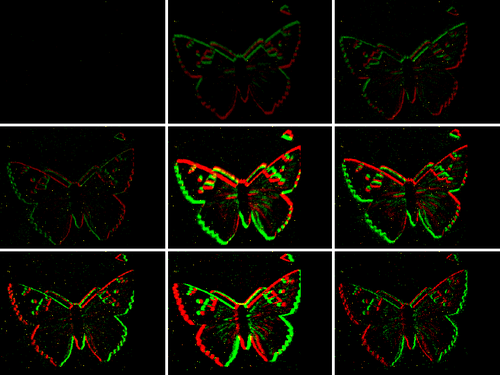}&
    	\includegraphics[height=1.6in]{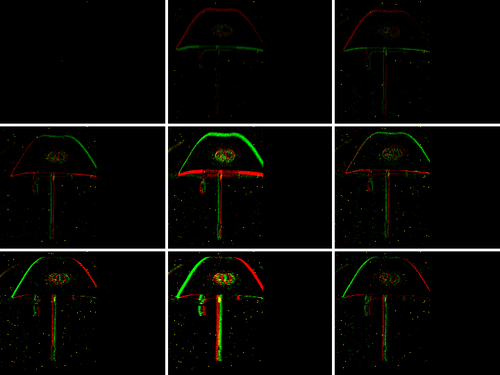}&
    	\includegraphics[height=1.6in]{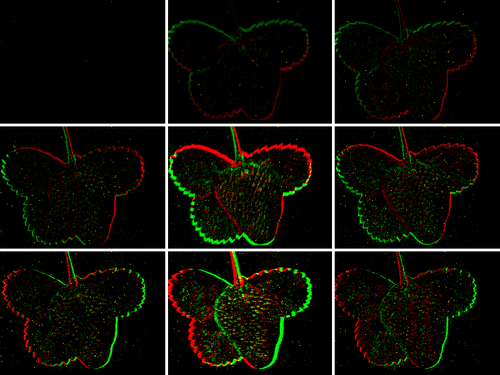}\\
        \includegraphics[height=1.3in]{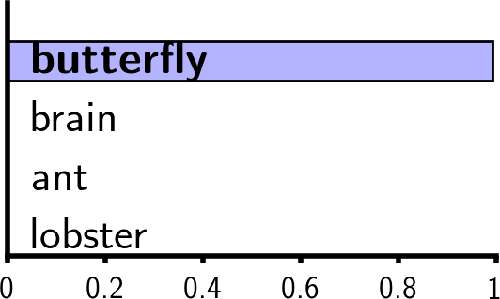}&
    	\includegraphics[height=1.3in]{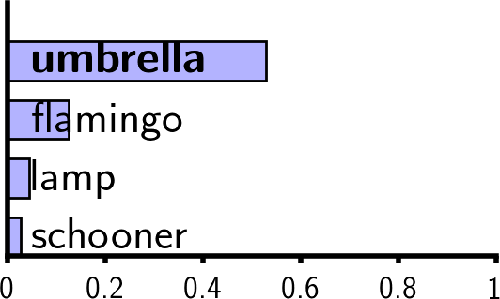}&
    	\includegraphics[height=1.3in]{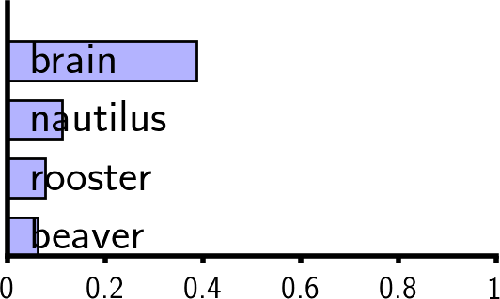}\\
        \\
    	(a)&(b)&(c)
    \end{tabular}
    \caption{Visualization of the event spike tensor (EST) representations derived from samples from the N-Caltech101 dataset \cite{Orchard15fns}. 
    The EST consists of 18 channels, where the first nine are filled with events of positive polarity and the last 9 are filled with negative polarity.  The figures show the nine temporal bins of the tensor with positive events in red and negative events in green. 
    We see that compared to N-Cars \cite{Sironi18cvpr} the event stream of this dataset is much cleaner and with much less noise.
    This is because the dataset was recorded in a controlled environment, by positioning an event camera toward an image projected on a screen.
    (a) and (b) correspond to correct predictions and (c) an incorrect one.}
    \label{fig:app:caltech_examples}
\end{figure*}

For the classification task, we investigated the relation between the number of temporal discretization bins, $B$, \emph{i.e.}, channels, of the event spike tensor (EST) and the network performance.
We quantitavively evaluated this effect on the N-Cars \cite{Sironi18cvpr} and N-Caltech101 \cite{Orchard15fns} datasets.
More specifically, we trained four networks, each using the learned EST with $B=2, 4, 9, 16$ and timestamp measurements, since this representation achieved the highest classification scores. 
The final input representations have $4$, $8$, $18$, and $32$ channels since we stack the polarity dimension along the temporal dimension.
\begin{table}[t]
\centering
\begin{tabular}{c|c|c}
\textbf{Temporal Bins} & N-Cars & N-Caltech101 \\ \hline
2 & 0.908 & 0.792 \\
4 & 0.912 & 0.816 \\
9 & \textbf{0.925} & 0.817 \\
16 & 0.923 & \textbf{0.837} \\
\end{tabular}
\caption{Classification accuracy on N-Cars \cite{Sironi18cvpr} and N-Caltech101 \cite{Orchard15fns} for input representations based on the event spike tensor (EST).
Four variations of the EST were tested, varying the number of temporal bins between $2,4,9$ and $16$.
The best representations are highlighted in bold.}
\label{tab:app:classification}
\end{table}
The results for this experiment are summarized in Tab.~1 and example classifications for the N-Cars and N-Caltech101 dataset are provided in Figs.~\ref{fig:app:cars_examples} and \ref{fig:app:caltech_examples}.

For both datasets, we observe a very similar trend in the dependency of classification accuracy to temporal discretization: performance appears to increase with finer discretization, \emph{i.e.}, with a larger number of channels.
However, for the N-Cars dataset performance plateaus after $B=9$ channels, while for the N-Caltech dataset performance continues to increase with a larger number of channels.
This difference  can be explained by the different qualities of the datasets.
While the N-Cars dataset features samples taken in an outdoor environment (Fig.~\ref{fig:app:cars_examples}), the N-Caltech101 samples were taken in controlled, constant lighting conditions and with consistent camera motion. 
This leads to higher quality samples in the N-Caltech101 dataset (Fig.~\ref{fig:app:caltech_examples}), while the samples in N-Cars are frequently corrupted by noise (Fig.~\ref{fig:app:cars_examples} (a-d)).
In low noise conditions (Fig.~\ref{fig:app:cars_examples} (a)) classification accuracy is very high ($99\%$).
However, as the signal decreases due to the lack of motions (Fig.~\ref{fig:app:cars_examples} (b-d)) the classification accuracy decreases rapidly.
Increasing the number of temporal bins further dilutes the signal present in the event stream, resulting in noisy channels (Fig.~\ref{fig:app:cars_examples} (c)), which impacts performance negatively.
In addition, more input channels results in higher the memory and computational costs of the network.
Therefore to trading-off performance for computational accuracy, we use $B=9$ in all our classification experiments.

\subsubsection{Optical Flow}
\label{sec:app:optical_flow}
\useunder{\uline}{\ul}{}
\begin{table*}[]
\centering
\begin{tabular}{c|c|c|cc|cc|cc}
\hline
\multirow{2}{*}{\textbf{Representation}} & \multirow{2}{*}{\textbf{Measurement}} & \multirow{2}{*}{\textbf{Kernel}} & \multicolumn{2}{c|}{indoor\_flying1} & \multicolumn{2}{c|}{indoor\_flying2} & \multicolumn{2}{c|}{indoor\_flying3} \\ \cline{4-9} 
                                         &                                       &                                  & AEE               & \% Outlier       & AEE               & \% Outlier       & AEE               & \% Outlier       \\ \hline
Event Frame                              & \multirow{4}{*}{polarity}             & \multirow{4}{*}{trilinear}       & 1.21              & 4.19             & 2.04              & 20.6             & 1.83              & 16.6             \\
Two-Channel Image                        &                                       &                                  & 1.31              & 4.75             & 2.05              & 23.2             & 1.83              & 11.4             \\
Voxel Grid                               &                                       &                                  & \textbf{0.96}     & 1.47             & 1.65              & 14.6             & 1.45              & 11.4             \\
\textbf{EST (Ours)}                      &                                       &                                  & 1.01              & 1.59             & 1.79              & 16.7             & 1.57              & 13.8             \\ \hline
Event Frame                              & \multirow{4}{*}{count}                & \multirow{4}{*}{trilinear}       & 1.25              & 3.91             & 2.11              & 22.9             & 1.85              & 17.1             \\
Two-Channel Image                        &                                       &                                  & 1.21              & 4.49             & 2.03              & 22.8             & 1.84              & 17.7             \\
Voxel Grid                               &                                       &                                  & 0.97              & 1.33             & 1.66              & 14.7             & 1.46              & 12.1             \\
\textbf{EST (Ours)}                      &                                       &                                  & 1.03              & 2.00             & 1.78              & 16.5             & 1.56              & 13.2             \\ \hline
Event Frame                              & \multirow{4}{*}{time stamps}          & \multirow{4}{*}{trilinear}       & 1.17              & 2.44             & 1.93              & 18.9             & 1.74              & 15.6             \\
Two-Channel Image                        &                                       &                                  & 1.17              & 1.50             & 1.97              & 14.9             & 1.78              & 11.7             \\
Voxel Grid                               &                                       &                                  & 0.98              & 1.20             & 1.70              & 14.3             & 1.50              & 12.0             \\
\textbf{EST (Ours)}                      &                                       &                                  & 1.00              & 1.35             & 1.71              & 11.4             & 1.51              & 8.29             \\ \hline
\multirow{3}{*}{\textbf{EST (Ours)}}     & \multirow{3}{*}{time stamps}          & alpha                            & 1.03              & 1.34             & 1.52              & 11.7             & 1.41              & 8.32             \\
                                         &                                       & exponential                      & \textbf{0.96}     & 1.27             & 1.58              & 10.5             & \textbf{1.40}     & 9.44             \\
                                         &                                       & learnt                           & 0.97              & \textbf{0.91}    & \textbf{1.38}     & \textbf{8.20}    & 1.43              & \textbf{6.47}    \\ \hline
\end{tabular}

\caption{Average end-point error (AEE) and \% of outliers evaluation on the MVSEC datasets. Ablation of different measurement functions for the event spike tensor. The best candidates are highlighted in bold.}
\label{tab:app:optical_flow_measurement}
\end{table*}

\begin{table*}[!htbp]
\centering
\begin{tabular}{c|cc|cc|cc}
\multirow{2}{*}{\textbf{Temporal Bins}} & \multicolumn{2}{c|}{\textit{indoor\_flying1}} & \multicolumn{2}{c|}{\textit{indoor\_flying2}} & \multicolumn{2}{c}{\textit{indoor\_flying3}} \\ \cline{2-7} 
                                   & AEE                   & \% Outlier            & AEE                   & \% Outlier            & AEE                   & \% Outlier           \\ \hline
2                                  & 0.97                  & 0.98 & 1.45                  & 8.86                  & 1.37                  & 6.66                 \\
4                                  & \textbf{0.96}         & 1.13         & 1.42         & 8.86         & 1.35         & \textbf{5.98}        \\
9                                 & 0.97                  & \textbf{0.91}         & \textbf{1.38}         & \textbf{8.20}       & 1.43                  & 6.47                 \\
16                                 & 0.95                  & 1.56                   & 1.39                  & 8.58                  & \textbf{1.34}                  & 6.82                
\end{tabular}%
\caption{Average end-point error (AEE) and \% of outliers for optical flow predictions on the MVSEC dataset \cite{Zhu18rss}. Four event representations based on the voxel grid were tested with 2, 4, 9 and 16 temporal bins. The best representation is highlighted in bold.}
\label{tab:app:optic_flow_table}
\end{table*}
In this section, we ablate two features of the representations used for optical flow prediction: (i) the measurement function $f$ (defined in \eqref{eq:spike_train_general}), and (ii) the number of temporal discretization bins, $B$.
We use the Multi Vehicle Stereo Event Camera (MVSEC) dataset \cite{Zhu18rss} for quantitative evaluation.

Tab.~\ref{tab:app:optical_flow_measurement} shows the performance of our candidate measurement functions, \emph{i.e.},\  \emph{polarity}, \emph{event count}, and event \emph{timestamp}, for the generation of the representations (see  \eqref{eq:spike_train_general_convolved}.
While it would be possible to learn the measurement function together with the kernel, in our experiments we have considered this function to be fixed.
This heuristic proved to speed-up convergence of our models, while decreasing the computational costs at training and inference time.

In Tab.~\ref{tab:app:optical_flow_measurement} it can be observed that the event timestamp yields the highest accuracy among the measurement functions.
This is indeed very intuitive since, while polarity and event count information is contained in the EST, the timestamp information is partially lost due to discretization.
Adding it back in the measurements gives the EST the least amount of information lost with respect to the original event point set, therefore maximizing the performance of end-to-end learning.

To understand the role that the number of temporal bins plays, we choose the best event representation for this task, the EST with timestamp measurements, and vary the number of temporal bins from $B=2,4,9,16$. 
The average endpoint errors and outlier ratios are reported in Tab.~\ref{tab:app:optic_flow_table}. 

As with the classification task (Sec.~\ref{sec:app:classification}), we observe a trade-off between using too few channels and too many.
Since MVSEC records natural outdoor scenes, event measurements are corrupted by significant noise.
As we increase the number of channels, the signal-to-noise ratio in the individual channels drops, leading to less accurate optical flow estimates.
In contrast, decreasing the number of channels also has adverse effects, as this removes valuable information from the event stream due to temporal aliasing effects.
Therefore, a compromise must be made between high and low channel numbers.
In the experiments reported in the paper we chose a channel number of nine, as this presents a good compromise.

In conclusion, we encourage the reader to watch the supplementary video to see the qualitative results of our method on optical flow prediction.
We have observed that, despite the application environment and illumination conditions, our method generates predictions which are not only accurate, but also temporally consistent without any postprocessing.


\clearpage
\clearpage
{\small
\bibliographystyle{ieee_fullname}
\bibliography{main.bbl}
}

\end{document}